\documentclass{sig-alternate-05-2015}
\usepackage[finalizecache=false,frozencache=true,cachedir=.]{minted}
\usepackage{varioref}
\usepackage[colorlinks=true,linkcolor=red]{hyperref}
\usepackage{cleveref}

\newcommand{\sphinxcode}[1]{\texttt{#1}}
\newcommand{\sphinxtitleref}[1]{\texttt{#1}}

\newcommand{\capstart}{}
\newcommand{\sphinxincludegraphics}[1]{\includegraphics[width=1\linewidth]{#1}}

\begin{document}






%

\title{Akid: A Library for Neural Network Research and Production From A Dataism Approach}
%
%
%
%
%

\numberofauthors{1} 
%
\author{
%
%
\alignauthor
  Shuai Li\\
  \affaddr Chinese University of Hong Kong
  \email lishuai918@gmail.com
}

\maketitle
\begin{abstract}
    Neural networks are a revolutionary but immature technique that is fast
evolving and heavily relies on data. To benefit from the newest development and
newly available data, we want the gap between research and production as small
as possibly. On the other hand, differing from traditional machine learning
models, neural network is not just yet another statistic model, but a model for
the natural processing engine --- the brain. In this work, we describe a neural network library named {\texttt
  akid}. It provides higher level of abstraction for entities (abstracted as
  blocks) in nature
upon the abstraction done on signals (abstracted as tensors) by Tensorflow,
characterizing the dataism observation that all entities in nature processes
input and emit out in some ways. It includes a full stack of software that provides
abstraction to let researchers focus on research instead of implementation, while at the same time the developed program
can also be put into production seamlessly in a distributed
environment, and be production ready.
At the top application stack, it provides out-of-box tools
for neural network applications. Lower down, akid provides a programming
paradigm that lets user easily build customized models. The distributed
computing stack handles
the concurrency and communication, thus letting models be
trained or deployed to a single GPU, multiple GPUs, or a
distributed environment without affecting how a model is
specified in the programming paradigm stack. Lastly, the
distributed deployment stack handles how the distributed
computing is deployed, thus decoupling the research prototype environment with the actual production environment,
and is able to dynamically allocate computing resources, so
development (Devs) and operations (Ops) could be separated.
It has been open source, and please refer to
http://akid.readthedocs.io/en/latest/ for documentation.
\end{abstract}

%
%

%
%

%
%



\section{Introduction}
\label{sec:introduction}

Neural network, which is a cornerstone technique of a pool of techniques
under the name of Deep Learning nowadays, seems to have
the potential to lead to another technology revolution. It has incurred wide
enthusiasm in industry, and serious consideration in public sector and impact
evaluation in government. However, though being a remarkable breakthrough in
high dimensional perception problems academically and intellectually
stimulating and promising \cite{Krizhevsky2012} \cite{Xiong2016} \cite{He}
\cite{Mallat2016} \cite{Zeiler2014}, it is still rather an immature technique that is
fast moving and in short of understanding \cite{Szegedy2013}. Temporarily its true value lies
in the capability to solve perception related data analytic problems in industry,
e.g. self-driving cars, detection of lung cancer etc. On the other hand, Neural
Network is a technique that heavily relies on a large volume of data. It is
critical for businesses that use such a technique to leverage on newly
available data as soon as possible, which helps form a positive feedback loop
that reinforces the quality of service.

Accordingly, to benefit from the newest development and newly available data,
we want the gap between research and production as small as possible. In this
package, we explore technology stack abstraction that enable fast research
prototyping and are production ready.

{\texttt akid} tries to provide a full stack of software that provides abstraction to
let researchers focus on research instead of implementation, while at the same
time the developed program can also be put into production seamlessly in a
distributed environment, and be production ready.

\begin{figure}
  \centering
  \includegraphics[width=0.6\linewidth]{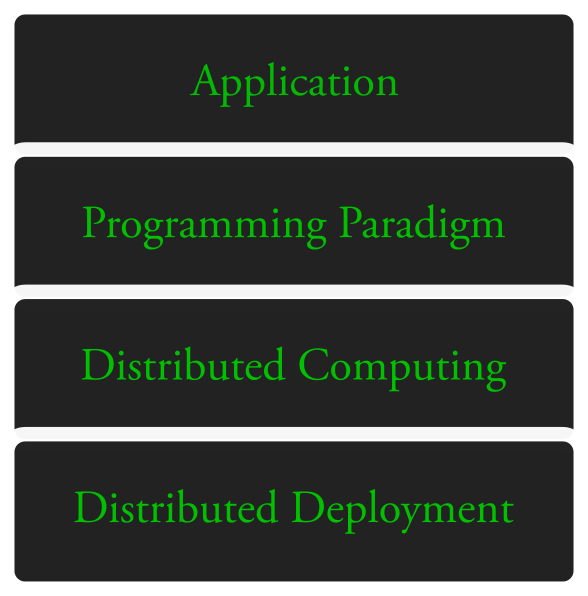}
  \caption{Illustration of stack abstraction of {\texttt akid}.}
  \label{fig:stack}
\end{figure}

At the top application stack, it provides out-of-box tools for neural network
applications. Lower down, {\texttt akid} provides programming paradigm that lets users
easily build customized models, which is the major intellectual innovation of
{\tt akid} that provides higher level of abstraction for entities in nature
(abstracted as blocks)
upon the abstraction done on signals (abstracted as tensors) by Tensorflow. The distributed computing stack handles the
concurrency and communication, thus letting models be trained or deployed to a
single GPU, multiple GPUs, or a distributed environment without affecting how a
model is specified in the programming paradigm stack. Lastly, the distributed
deployment stack handles how the distributed computing is deployed, thus
decoupling the research prototype environment with the actual production
environment, and is able to dynamically allocate computing resources, so
development (Devs) and operations (Ops) could be separated. An illustration of
the four stack is shown in \Cref{fig:stack}.

From a feature point of view as a library, it aims to enable fast prototyping
and production ready at the same time by offering the following features:
\begin{itemize}
\item {} 
supports fast prototyping
\begin{itemize}
\item {} 
built-in data pipeline framework that standardizes data preparation and
data augmentation.

\item {} 
arbitrary connectivity schemes (including multi-input and multi-output
training), and easy retrieval of parameters and data in the network

\item {} 
meta-syntax to generate neural network structure before training

\item {} 
support for visualization of computation graph, weight filters, feature
maps, and training dynamics statistics.

\end{itemize}

\item {} 
be production ready
\begin{itemize}
\item {} 
built-in support for distributed computing

\item {} 
compatibility to orchestrate with distributed file systems, docker
containers, and distributed operating systems such as Kubernetes.

\end{itemize}

\end{itemize}

The name comes from the Kid saved by Neo in \emph{Matrix}, and the metaphor to
build a learning agent, which we call \emph{kid} in human culture.

The rest of the paper discusses related works in \Cref{sec:related-works},
each stack of {\texttt akid} in detail in \Cref{intros/index:akid-stack}.

\section{Related works}
\label{sec:related-works}

\sphinxcode{akid} differs from existing packages from the perspective that it
does not aim to be yet another wrapper for another machine learning
model. Subtle it seems. The fundamental difference lies in the design. It aims to
reproduce how signal propagates in nature by introducing {\it Block}. If Tensor
in Tensorflow can be viewed as the abstraction for signals in nature, Block can
be viewed as the abstraction for entities in nature, which all process inputs
in some way, and emit output. It also aims to integrate technology stacks to
solve both research prototyping and industrial production by clearly defining
the behavior for each stack. We compare \sphinxcode{akid} with existing
packages in the following briefly.  Note that since Tensorflow is used as the
computation backend, we do not discuss speed here, which is not our concern for
{\texttt akid}.

Theano \cite{TheTheanoDevelopmentTeam2016}, Torch \cite{Collobert2002},
Caffe \cite{Jia2014}, MXNet \cite{Chen2016} are packages that aim to provide a
friendly front end to complex computation back-end that are written in
C++. Theano is a python front end to a computational graph compiler, which has
been largely superseded by Tensorflow in the compilation speed, flexibility,
portability etc, while \sphinxcode{akid} is built on of Tensorflow. MXNet is a competitive
competitor to Tensorflow. Torch is similar with theano, but with the front-end
language to be Lua, the choice of which is mostly motivated from the fact that
it is much easier to interface with C using Lua than Python. It has been widely
used before deep learning has reached wide popularity, but is mostly a quick
solution to do research in neural networks when the integration with community
and general purpose production programming are not pressing. Caffe is written
in C++, whose friendly front-end, aka the text network configuration file,
loses its affinity when the model goes more than dozens of
layer.

\href{https://deeplearning4j.org/}{DeepLearning4J} is an industrial solution
to neural networks written in Java and Scala, and is too heavy weight for
research prototyping.

Perhaps the most similar package existing with \sphinxcode{akid} is
\href{http://akid.readthedocs.io/en/latest/intros/keras.io}{Keras}, which both aim to provide a more intuitive interface to
relatively low-level library, i.e. Tensorflow. \sphinxcode{akid} is different from Keras in
at least two fundamental aspects. First, \sphinxcode{akid} mimics how signals propagate
in nature by abstracting everything as a semantic block, which holds many
states, thus is able to provide a wide range of functionalities in a easily
customizable way, while Keras uses a functional API that directly manipulates
tensors, which is a lower level of abstraction, e.g. it has to do class
attributes traverse to retrieve layer weights with a fixed variable name while
in \sphinxcode{akid} variable are retrieved by names. Second, Keras mostly only provides
an abstraction to build a neural network topology, which is roughly the
programming paradigm stack of \sphinxcode{akid}, while \sphinxcode{akid} provides unified abstraction
that includes application stack, programming stack, and distributed computing
stack. A noticeable improvement is that Keras needs the user to handle communication
and concurrency, while the distributed computing stack of \sphinxcode{akid} hides them.

\section{\texttt{akid} stack}
\label{intros/index:akid-stack}

Now we go technical to discuss each stack provided by \sphinxcode{akid}. The
major novel intellectual design of \sphinxcode{akid} is the programming
paradigm that provides a higher level abstraction upon signal/tensor. We
introduce it first, then we discuss the application stack, and distributed
computing and deployment stack.

\subsection{Programming Paradigm}
\label{intros/index:programming-paradigm}

\subsubsection{It is all about signal processing blocks}
\label{sec:it-all-about}

\begin{figure}[htb]
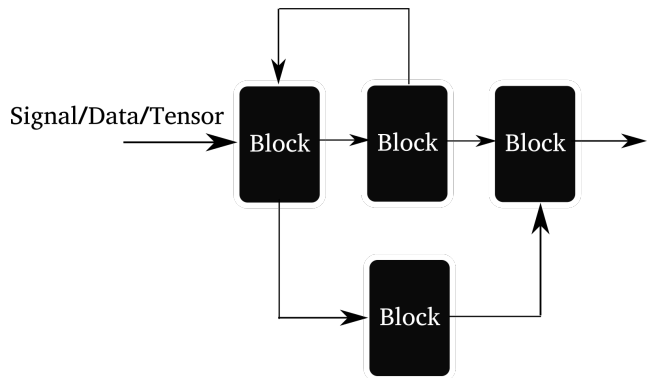

\centering
\capstart

\noindent\sphinxincludegraphics{./{akid_block}.png}
\caption{Illustration of the arbitrary connectivity supported by \sphinxtitleref{akid}. Forward
connection, branching and mergine, and feedback connection are supported.}\label{intros/index:id5}\end{figure}
\phantomsection\label{intros/index:module-akid.core.blocks}\index{akid.core.blocks (module)}

\sphinxtitleref{akid} builds another layer of abstraction on top of \emph{Tensor}: \emph{Block}.
Tensor can be taken as the media/formalism signal propagates in digital world,
while Block is the data processing entity that processes inputs and emits
outputs.

It coincides with a branch of ``ideology'' called dataism that takes everything
in this world is a data processing entity. An interesting one that may come
from \emph{A Brief History of Tomorrow} by Yuval Noah Harari.

Best designs mimic nature. \sphinxtitleref{akid} tries to reproduce how signals in nature
propagates. Information flow can be abstracted as data propagating through
inter-connected blocks, each of which processes inputs and emits outputs. For
example, a vision classification system is a block that takes image inputs and
gives classification results. Everything is a \sphinxtitleref{Block} in \sphinxtitleref{akid}.

A block could be as simple as a convonlutional neural network layer that merely
does convolution on the input data and outputs the results; it also be as
complex as an acyclic graph that inter-connects blocks to build a neural
network, or sequentially linked block system that does data augmentation.

Compared with pure symbol computation approach, like the one in tensorflow, a
block is able to contain states associated with this processing unit. Signals
are passed between blocks in form of tensors or list of tensors. Many heavy
lifting has been done in the block (\sphinxtitleref{Block} and its sub-classes),
e.g. pre-condition setup, name scope maintenance, copy functionality for
validation and copy functionality for distributed replicas, setting up and
gathering visualization summaries, centralization of variable allocation,
attaching debugging ops now and then etc.

\sphinxcode{akid} offers various kinds of blocks that are able to connect to other blocks
in an arbitrary way, as illustrated in \Cref{intros/index:id5}. It is also easy to build one's own
blocks. The \sphinxcode{Kid} class is essentially an assembler that assemblies blocks
provided by \sphinxcode{akid} to mainly fulfill the task to train neural networks. Here we
show how to build an arbitrary acyclic graph of blocks using class {\texttt Brain}, to illustrate how to
use blocks in \sphinxcode{akid}.
\phantomsection\label{intros/index:module-akid.core.brains}\index{akid.core.brains (module)}

A brain is the data processing engine to process data supplied by \sphinxtitleref{Sensor} to
fulfill certain tasks. More specifically,
\begin{itemize}
\item {} 
it builds up blocks to form an arbitrary network

\item {} 
offers sub-graphs for inference, loss, evaluation, summaries

\item {} 
provides access to all data and parameters within

\end{itemize}

To use a brain, data as a list should be fed in, as how it is done in with any other
block. Some pre-specified brains are available under \sphinxtitleref{akid.models.brains}. An
example could be:

\begin{minted}{python}
# ... first get a feed sensor
sensor.setup()
brain = OneLayerBrain(name="brain")
input = [sensor.data(), sensor.labels()]
brain.setup(input)
\end{minted}

Note in this case, \sphinxtitleref{data()} and \sphinxtitleref{labels()} of \sphinxtitleref{sensor} returns tensors. It is
not always the case. If it does not, saying return a list of tensors, you need
do things like:

\begin{minted}{python}
input = [sensor.data()]
input.extend(sensor.labels())
\end{minted}

Act accordingly.

Similarly, all blocks work this way.

A brain provides easy ways to connect blocks. For example, a one layer brain
can be built through the following:

\begin{minted}{python}
  class OneLayerBrain(Brain):
    def __init__(self, **kwargs):
        super(OneLayerBrain, self).__init__(**kwargs)
        self.attach(
            ConvolutionLayer(ksize=[5, 5],
                            strides=[1, 1, 1, 1],
                            padding="SAME",
                            out_channel_num=32,
                            name="conv1")
        )
        self.attach(ReLULayer(name="relu1"))
        self.attach(
            PoolingLayer(ksize=[1, 5, 5, 1],
                        strides=[1, 5, 5, 1],
                        padding="SAME",
                        name="pool1")
        )

        self.attach(InnerProductLayer(
            out_channel_num=10, name="ip1"))
        self.attach(SoftmaxWithLossLayer(
            class_num=10,
            inputs=[
                {"name": "ip1", "idxs": [0]},
                {"name": "system_in", "idxs": [1]}],
            name="loss"))
\end{minted}

It assembles a convolution layer, a ReLU Layer, a pooling layer, an inner
product layer and a loss layer. To attach a block (layer) that directly takes
the outputs of the previous attached layer as inputs, just directly attach the
block. If \sphinxtitleref{inputs} exists, the brain will fetch corresponding tensors by name
of the block attached and indices of the outputs of that layer. See the loss
layer above for an example. Note that even though there are multiple inputs for
the brain, the first attached layer of the brain will take the first of these
input by default, given the convention that the first tensor is the data, and
the remaining tensors are normally labels, which is not used till very late.

\begin{figure}[htb]
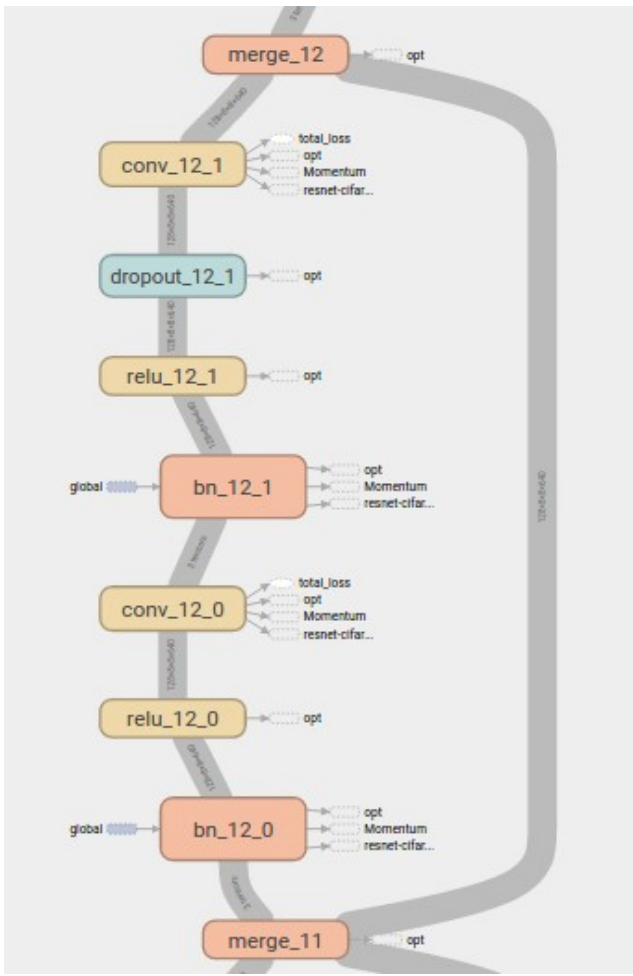

\centering
\capstart

\noindent\sphinxincludegraphics{./{residual_block}.png}
\caption{A residual unit. On the left is the branch that builds up patterns
complexity, and on the right is the stem branch that shortcuts any layers to
any layers. They merge at the at the start and at the end of the branching
points.}\label{intros/index:id6}\end{figure}

As an example to build more complex connectivity scheme, residual units can be
built using \sphinxcode{Brain} as shown in \Cref{intros/index:id6}.

\subsubsection{Self-modifying brains --- parameter tuning}
\label{intros/index:parameter-tuning}

\sphinxcode{akid} offers automatic parameter tuning through defining template using \sphinxcode{tune}
function. A function \sphinxtitleref{tune} that takes a Brain jinja2 template class and a parameters
to fill the template in runtime. 

The \sphinxtitleref{tune} function would use all available GPUs to train
networks with all given different set of parameters. If available GPUs are not
enough, the ones that cannot be trained will wait till some others finish, and
get its turn.

Tunable parameters are divided into two sets, network hyper parameters,
\sphinxtitleref{net\_paras\_list}, and optimization hyper parameters, \sphinxtitleref{opt\_paras\_list}. Each
set is specified by a list whose item is a dictionary that holds the actual
value of whatever hyper parameters defined as jinja2 templates. Each item
in the list corresponds to a tentative training instance. Network paras and
optimization paras combine with each other exponentially (or in Cartesian
Product way if we could use Math terminology), which is to say if you have
two items in network parameter list, and two in optimization parameters,
the total number of training instances will be four.

Given the available GPU numbers, a semaphore is created to control access
to GPUs. A lock is created to control access to the mask to indicator which
GPU is available. After a process has modified the gpu mask, it releases
the lock immediately, so other process could access it. But the semaphore
is still not release, since it is used to control access to the actual
GPU. A training instance will be launched in a subshell using the GPU
acquired. The semaphore is only released after the training has finished.

For example, to tune the activation function and learning rates of a
network, first we set up network parameters in \sphinxtitleref{net\_paras\_list},
optimization parameters in \sphinxtitleref{opt\_paras\_list}, build a network in the \sphinxtitleref{setup}
function, then pass all of it to tune:

\begin{minted}{python}
  net_paras_list = []
net_paras_list.append({
    "activation": [
        {"type": "relu"},
        {"type": "relu"},
        {"type": "relu"},
        {"type": "relu"}],
    "bn": True})
net_paras_list.append({
    "activation": [
        {"type": "maxout", "group_size": 2},
        {"type": "maxout", "group_size": 2},
        {"type": "maxout", "group_size": 2},
        {"type": "maxout", "group_size": 5}],
    "bn": True})

opt_paras_list = []
opt_paras_list.append({"lr": 0.025})
opt_paras_list.append({"lr": 0.05})

def setup(graph):

    brain.attach(cnn_block(
        ksize=[8, 8],
        init_para={
            "name": "uniform",
            "range": 0.005},
        wd={"type": "l2", "scale": 0.0005},
        out_channel_num=384,
        pool_size=[4, 4],
        pool_stride=[2, 2],
        activation={{ net_paras["activation"][1] }},
        keep_prob=0.5,
        bn={{ net_paras["bn"] }}))

tune(setup, opt_paras_list, net_paras_list)
\end{minted}

\subsection{Application stack}
\label{intros/index:application-stack}
At the top of the stack, \sphinxcode{akid} could be used as a part of application without
knowing the underlying mechanism of neural networks.

\sphinxcode{akid} provides full machinery from preparing data, augmenting data, specifying
computation graph (neural network architecture), choosing optimization
algorithms, specifying parallel training scheme (data parallelism etc), logging
and visualization.

\subsubsection{Neural network training --- A holistic example}
\label{intros/index:neural-network-training-a-holistic-example}
To create better tools to train neural network has been at the core of the
original motivation of \sphinxcode{akid}. Consequently, in this section, we describe how
\sphinxcode{akid} can be used to train neural networks. Currently, all the other features
resolve around this.

The snippet below builds a simple neural network, and trains it using MNIST,
the digit recognition dataset.

\begin{minted}{python}
from akid import AKID_DATA_PATH
from akid import FeedSensor
from akid import Kid
from akid import MomentumKongFu
from akid import MNISTFeedSource

from akid.models.brains import LeNet

brain = LeNet(name="Brain")
source = MNISTFeedSource(
    name="Source",
    url='http://yann.lecun.com/exdb/mnist/',
    work_dir=AKID_DATA_PATH + '/mnist',
    center=True,
    scale=True,
    num_train=50000,
    num_val=10000)

sensor = FeedSensor(name='Sensor', source_in=source)
s = Kid(sensor,
        brain,
        MomentumKongFu(name="Kongfu"),
        max_steps=100)
kid.setup()
kid.practice()
\end{minted}

\begin{figure}
  \centering
  \includegraphics[width=0.6\linewidth]{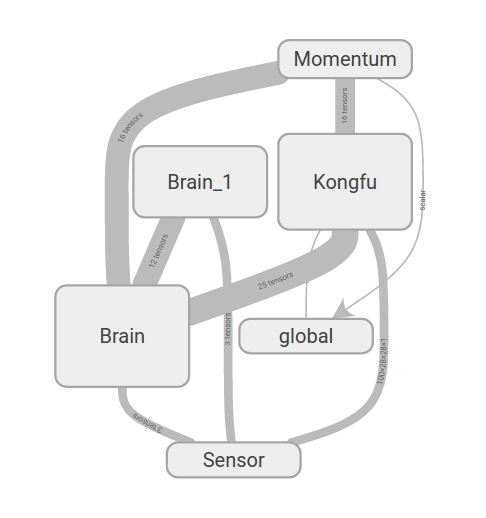}
  \caption{Computational graph of the simple neural network built for MNIST
    digit recognition example.}
  \label{fig:mnist_graph}
\end{figure}

It builds a computation graph as shown in \Cref{fig:mnist_graph}.

The underlying stories are described in the following section, which also
debriefs the design motivation and vision behind {\texttt akid}.

\sphinxcode{akid} is a kid who has the ability to keep practicing to improve itself. The
kid perceives a data \sphinxcode{Source} with its \sphinxcode{Sensor} and certain learning methods
(nicknamed \sphinxcode{KongFu}) to improve itself (its \sphinxcode{Brain}), to fulfill a certain
purpose. The world is timed by a clock. It represents how long the kid has been
practicing. Technically, the clock is the conventional training step.

To break things done, \sphinxcode{Sensor} takes a \sphinxcode{Source} which either provides data in
form of tensors from Tensorflow or numpy arrays. Optionally, it can make jokers
on the data using \sphinxcode{Joker}, meaning doing data augmentation. The data processing
engine, which is a deep neural network, is abstracted as a \sphinxcode{Brain}. \sphinxcode{Brain} is
the name we give to the data processing system in living beings. A \sphinxcode{Brain}
incarnates one of data processing system topology, or in the terminology of
neural network, network structure topology, such as a sequentially linked
together layers, to process data. Available topology is defined in module
\sphinxcode{systems}. The network training methods, which are first order iterative
optimization methods, is abstracted as a class \sphinxcode{KongFu}. A living being needs
to keep practicing Kong Fu to get better at tasks needed to survive.

A living being is abstracted as a \sphinxcode{Kid} class, which assemblies all above
classes together to play the game. The metaphor means by sensing more examples,
with certain genre of Kong Fu(different training algorithms and policies), the
data processing engine of the \sphinxcode{Kid}, the brain, should get better at doing
whatever task it is doing, letting it be image classification or something
else.

\subsubsection{Visualization}
\label{intros/index:visualization}
As a library gearing upon research, it also has rich features to visualize
various components of a neural network. It has built-in training dynamics
visualization, more specifically, distribution visualization on
multi-dimensional tensors, e.g., weights, activation, biases, gradients, etc,
and line graph visualization on on scalars, e.g., training loss, validation
loss, learning rate decay, regularization loss in each layer, sparsity of neuron
activation etc, and filter and feature map visualization for neural networks.

\begin{figure}[htb]
\centering

 \noindent\includegraphics[width=1\linewidth]{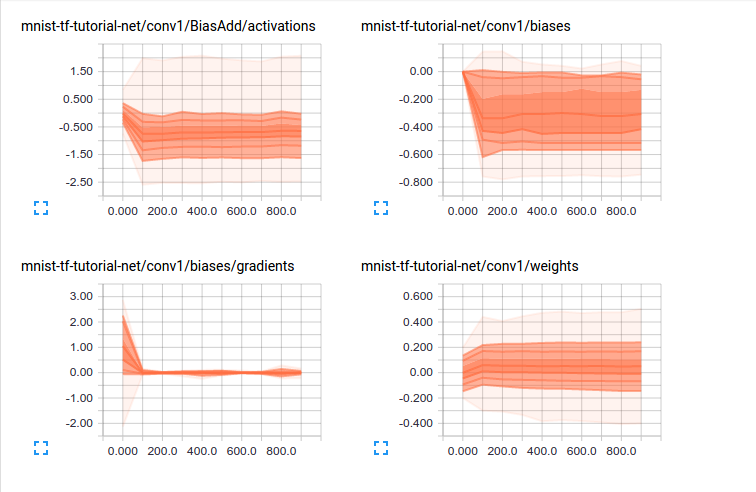}
\caption{Visualization of how distribution of multi-dimensional tensors change over
time.  Each line on the chart represents a percentile in the distribution
over the data: for example, the bottom line shows how the minimum value has
changed over time, and the line in the middle shows how the median has
changed. Reading from top to bottom, the lines have the following meaning:
{[}maximum, 93\%, 84\%, 69\%, 50\%, 31\%, 16\%, 7\%, minimum{]} These percentiles can
also be viewed as standard deviation boundaries on a normal distribution:
{[}maximum, \(\mu\)+1.5\(\sigma\), \(\mu\)+\(\sigma\), \(\mu\)+0.5\(\sigma\), \(\mu\), \(\mu\)-0.5\(\sigma\), \(\mu\)-\(\sigma\), \(\mu\)-1.5\(\sigma\), minimum{]} so that the
colored regions, read from inside to outside, have widths {[}\(\sigma\), 2\(\sigma\), 3\(\sigma\){]}
respectively.}\label{intros/index:id1}
\end{figure}

\begin{figure}[htb]
\centering

\noindent\includegraphics[width=1\linewidth]{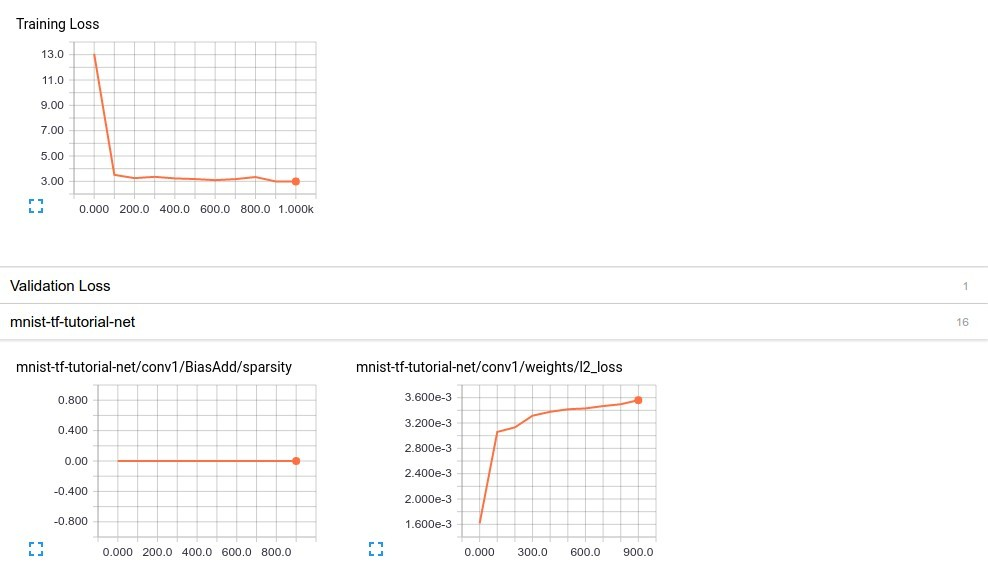}
  
\caption{Visualization of how important scalar measures change over time.}\label{intros/index:id2}\end{figure}

Distribution and scalar visualization are built in for typical parameters and
measures, and can be easily extended, and distributedly gathered. Distribution
visualizations are shown in \Cref{intros/index:id1}, and scalar
visualizations are shown in \Cref{intros/index:id2}.

\sphinxcode{akid} supports visualization of all feature maps and filters with control on
the layout through \sphinxcode{Observer} class. When having finished creating a \sphinxcode{Kid},
pass it to \sphinxcode{Observer}, and call visualization as the following.

\begin{minted}{python}
from akid import Observer

o = Observer(kid)
# Visualize filters as the following
o.visualize_filters()
# Or visualize feature maps as the following
o.visualize_activation()
\end{minted}

\begin{figure}[htb]
\centering

\noindent\includegraphics[width=1\linewidth]{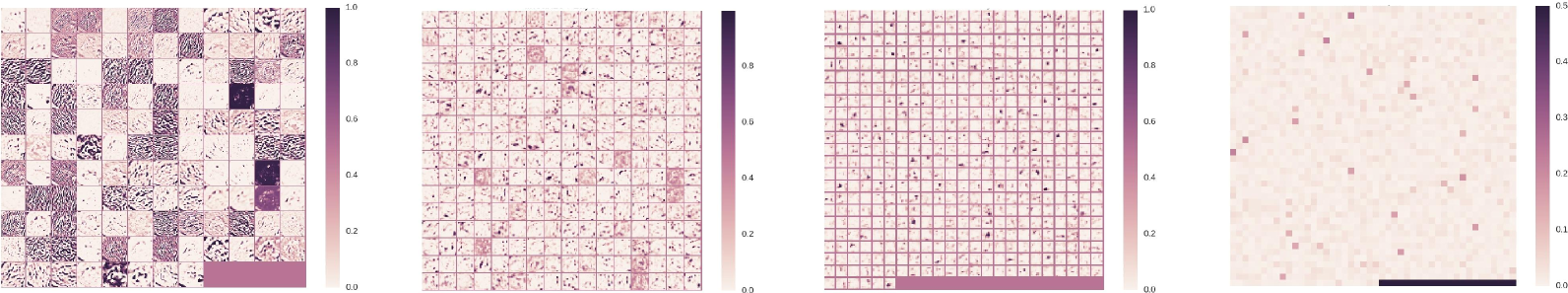}
\caption{Visualization of feature maps learned.}\label{intros/index:id3}\end{figure}

\begin{figure}[htb]
\centering

\noindent\includegraphics[width=1\linewidth]{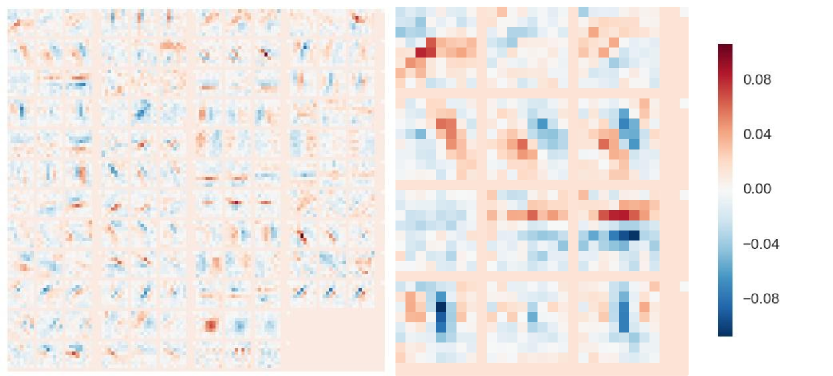}
\caption{Visualization of filters learned.}\label{intros/index:id4}\end{figure}

Various layouts are provided when drawing the filters. Additional features are
also available. The post-preprocessed visualization results of filters are
shown in \Cref{intros/index:id4}, and that of feature maps are shown
\Cref{intros/index:id3}.


\subsection{Distributed Computation}
\label{intros/index:distributed-computation}\label{intros/index:module-akid.core.engines}\index{akid.core.engines (module)}

\begin{figure*}[t]
\centering
\includegraphics[width=0.8\linewidth]{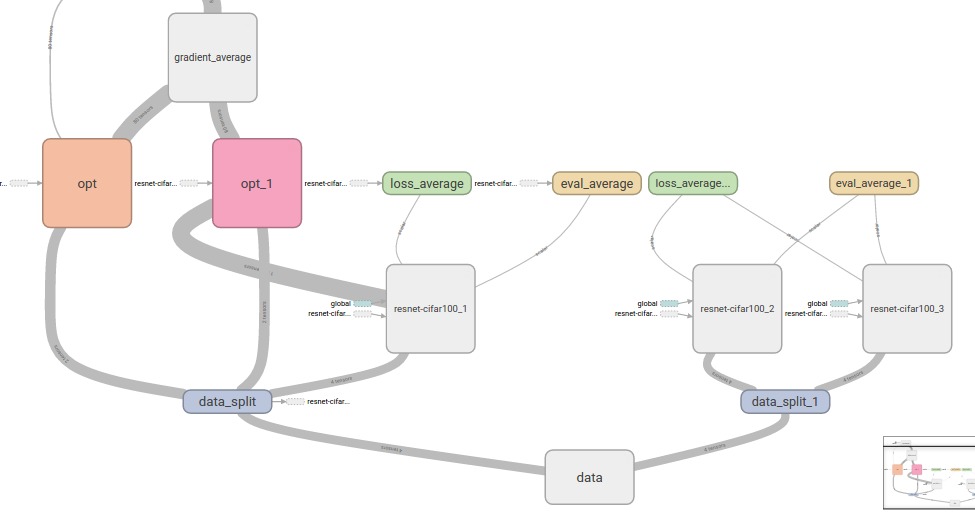}
\caption{Illustration of computational graph constructed by a data parallel engine.
It partitions a mini-batch of data into subsets, as indicated by the
\sphinxtitleref{data\_split} blue blocks, and passes the subsets to replicates of neural
network models at different computing towers, as indicated by the gray blocks
one level above blue blocks, then after the inference results have been
computed, the results and the labels (from the splitted data block) will be
passed to the optimizers in the same tower, as indicated by red and orange
blocks named \sphinxtitleref{opt}, to compute the gradients. Lastly, the gradients will be
passed to a tower that computes the average of the gradients, and pass them
back to neural networks of each computing towers to update their
parameters.}\label{intros/index:id7}
\end{figure*}

The distributed computing stack is responsible to handle concurrency and
communication between different computing nodes, so the end user only needs to
deal with how to build a power network. All complexity has been hidden in the
class \sphinxtitleref{Engine}. The usage of \sphinxtitleref{Engine} is just to pick and use.

More specifically, \sphinxtitleref{akid} offers built-in data parallel scheme in form of class
\sphinxtitleref{Engine}. Currently, the engine mainly works with neural network training,
which is be used with \sphinxtitleref{Kid} by specifying the engine at the construction of the
kid.

As an example, we could do data parallelism on multiple towers using:

\begin{minted}{python}
  kid = kids.Kid(
    sensor,
    brain,
    MomentumKongFu(
        lr_scheme={
          "name": LearningRateScheme.placeholder}),
    engine={"name": "data_parallel", "num_gpu": 2},
    log_dir="log",
    max_epoch=200)
\end{minted}

The end computational graph constructed is illustrated in \Cref{intros/index:id7}.

\subsection{Distributed Deployment}
\label{intros/index:distributed-deployment}

The distributed deployment stack handles the actual production environment,
thus decouples the development/prototyping environment and production
environment. We leverage on recent developments of distributed system
in the open source community to build a distributed deployment solution for
akid. More specifically, we investigate and test out three cornerstone techniques
that provides network file system, i.e. {\texttt Glusterfs}, containerization,
i.e. {\texttt Docker}, and distributed scheduler, i.e. {\texttt
Kubernetes}, functionality. More would come when we get the chance to test them
out in a real production environment.

\section{Conclusion}
\label{sec:conclusion}

We have described {\texttt akid}, a neural network library that provides a four-layer stack to
enable fast research prototyping and be production ready. It has a clean and
intuitive application facing interface, nature inspired programming paradigm,
and abstracts away distributing computing, decouples developments and
operations.

\newpage
\bibliography{../../mendeley.bib/library}
\bibliographystyle{abbrv}

\end{document}